\documentclass[letterpaper, 10 pt, conference]{ieeeconf}  

\IEEEoverridecommandlockouts                              

\usepackage{graphics} 
\usepackage{epsfig} 
\usepackage{amsmath} 

\usepackage{graphics} 
\usepackage{epsfig} 
\usepackage{mathptmx} 
\usepackage{times} 
\usepackage{amsmath} 
\usepackage{amssymb}  
\usepackage{subfigure}
\usepackage{algorithm} 
\usepackage{algpseudocode}
\usepackage{multirow} 

\title{\LARGE \bf
Conquering Ghosts: Relation Learning for Information Reliability Representation and End-to-End Robust Navigation
}

\author{Kefan Jin, Xingyao Han, Jiangyuan Zhao, Hongye Wang
\thanks{K. Jin is with the MoE Key Lab of Artificial Intelligence, AI Institute, Shanghai Jiao Tong University, China. X. Han and H. Wang are with the Department of Automation, Shanghai Jiao Tong University, China. J. Zhao is with the Department of Computer Science, Shanghai Jiao Tong
University, China.}
}

\begin{document}

\maketitle
\thispagestyle{empty}
\pagestyle{empty}

\begin{abstract}
Environmental disturbances, such as sensor data noises, various lighting conditions, challenging weathers and external adversarial perturbations, are inevitable in real self-driving applications. Existing researches and testings have shown that they can severely influence the vehicle's perception ability and performance, one of the main issue is the false positive detection, i.e., the ``ghost'' object which is not real existed or occurs in the wrong position (such as a non-existent vehicle). 
Traditional navigation methods tend to avoid every detected objects for safety, 
however, avoiding a ``ghost'' object may lead the vehicle into a even more dangerous situation, such as a sudden break on the highway. 
Considering the various disturbance types, it is difficult to address this issue at the perceptual aspect. A potential solution is to detect the ghost through relation learning among the whole scenario and develop an integrated end-to-end navigation system. 
Our underlying logic is that the behavior of all vehicles in the scene is influenced by their neighbors, and normal vehicles behave in a logical way, while ``ghost'' vehicles do not. By learning the spatio-temporal relation among surrounding vehicles, an information reliability representation is learned for each detected vehicle and then a robot navigation network is developed. In contrast to existing works, we encourage the network to learn how to represent the reliability and how to aggregate all the information with uncertainties by itself, thus increasing the efficiency and generalizability. 
To the best of the authors' knowledge, this paper provides the first work on using graph relation learning to achieve end-to-end robust navigation in the presence of ghost vehicles. Simulation results in the CARLA platform demonstrate the feasibility and effectiveness of the proposed method in various scenarios.

\end{abstract}

\begin{figure}[!t]
\centering
{\includegraphics[width=1\columnwidth]{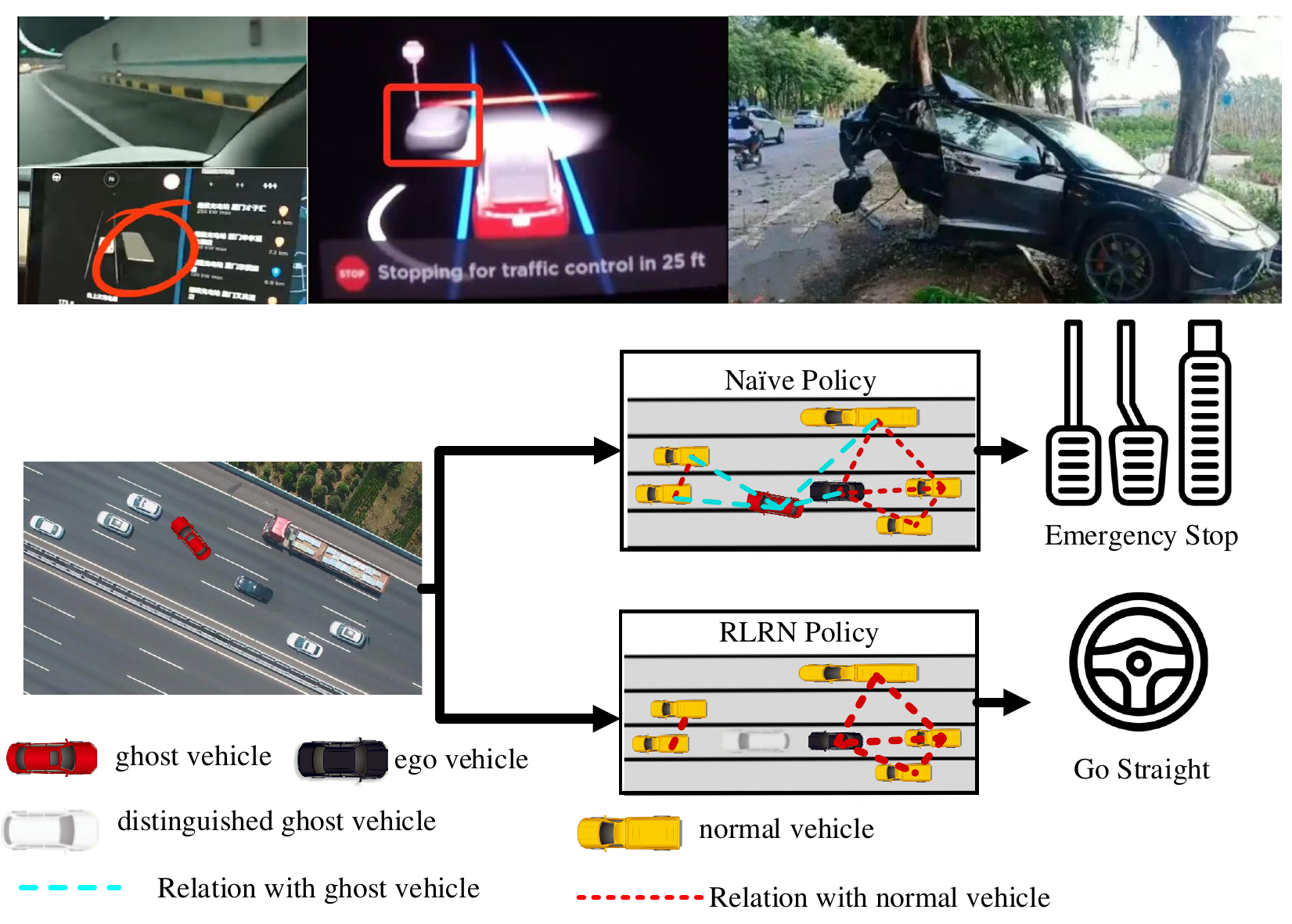}}
\caption{The real ghost cases in existing self-driving systems (upper) and our main idea (lower). When a ghost vehicle occurs, the navigation system (naive policy) will execute an emergency brake, which may lead to a collision with the vehicle behind. On the contrary, our proposed RLRN approach is able to distinguish the ghost vehicle with relation learning, thus eliminating the effect of ghost vehicles.}
\label{fig:Demo}
\end{figure}

\section{INTRODUCTION}
Researchers are increasingly focusing on the autonomous driving tasks 
in the recent decades \cite{deepDrive1,CIL}. An ideal navigation policy is expected to coordinate motion with surrounding dynamic objects, while reaching the destination according to its own mission requirements. The application of self-driving in practice is challenging due to the various complex environment and un-accurate perception of dynamic objects. 
Traditional approaches adopt modular paradigm which divides the whole system into several sub-modules \cite{pathPlanning,deepDrive1}, such as the environment perception, localisation, path planning, and maneuver control. However, without necessary feedback among each module, the error of each module will be accumulated and affect the performance of the following modules, which may lead to critical problems and require extra manufactured engineering effort. 
Recently, the rise of deep-learning methods provide advanced solutions \cite{deepDrive1,learningByCheating,CIL}. The end-to-end paradigm, as the most promising solution, considers the navigation system as a whole and directly maps the observation input to the final control action. 
Several researches aim to learn driving policy directly from the raw image data \cite{cnnNavi1} and LIDAR point cloud \cite{LIDAR}, however, their performance in the complex scenarios cannot be ensured, as it is very difficult to learn the semantic relations among objects in complex traffic scenario (which is very important for robust navigation). Since the semantic information can provide informative abstraction of the perceptual results and can be easily accessed, it is commonly utilized in existing end-to-end driving models to improve transferability, examples include the birdeye-views (BEVs) \cite{E2E1_semantic, E2E3_semantic} and high definition (HD) maps \cite{HD1}). Thanks to such semantic data and abstract vehicle states, the scalability to unseen scenarios can be easily achieved \cite{E2E3_semantic}. 



However, there are still several critical challenges which affect the practical application performance of the existing models. One critical issue is the false positive detection, i.e., the ghost object which is not real existed or occurs in the wrong position (such as a non-existent vehicle). As shown in the upper side of Fig.~\ref{fig:Demo}, the Tesla users have reported several cases that the perception system detects a non-existent vehicle and the self-driving car breaks suddenly, which further leads to dangerous situation and even crash accidents. 
The main reasons which lead to false positive detections can be summarized into three types: 1) Existing self-driving systems work well in semi-closed environments, their performance decreases largely 
in the presence of unseen conditions and unexpected environmental disturbances. For example, challenging weather conditions (such as fog and heavy rain), various illumination conditions and several corner cases (such as light reflection and vehicle-like figures in the traffic scenario) may decrease the detection and tracking performance, thus leading to mis-perception. 2) Current methods highly rely on deep-learning models, which can fail on the input data not well represented by the training dataset \cite{OOD}. It happens sometimes that a non-existent vehicle is mis-detected or an existing vehicle is tracked to a wrong position. 
3) Recently demonstrations have shown that crafted perturbations can easily fool the existing state-of-the-art autonomous driving systems \cite{lidarAtt1}. Most recent researches in the CV community also show that adversarial attacks with minimal disturbances can easily cause arbitrary perceptual misleadings \cite{lidarAtt2}. 
For a most recent example, on $17^{th}$ Feb. 2022, the National Highway Traffic Safety Administration announced that it would launch a formal investigation into 416,000 Tesla vehicles for unreasonable emergency braking resulting from such ghost vehicles \footnote{https://www.163.com/dy/article/H0I8C5CD05278GV4.html}.

Resolving the ghost issue is very challenging due to the complex external disturbances and various attacking methods. Existing robust methods are usually only effective on one specific interference and cannot resist unseen disturbances \cite{OOD}. In this paper, we do not explicitly detect the interference from the aspect of perception, instead, we aim to represent the reliability by using the relation learning and let the navigation model to learn how to aggregate the information with uncertainties by itself. 
In reality, the motion of all vehicles are influenced by their surrounding vehicles as well as the traffic scenario information, which means that there exist implicit spatio-temporal relations/consistencies among the behavior of all vehicles, and their behavior should also be consistent with the whole scenarios information. By learning this consistency, the ghost vehicle can be recognized. 
Therefore, compared with improving robustness of perception systems to various interference, distinguishing and excluding the influence of ghost vehicles in the navigation model can be a better and easier way. 

In this work, a Reliability Leaning based Robust Navigation (RLRN) approach is proposed to achieve robust navigation against ghost vehicles 
Our approach addresses the ghost issue by learning the behavior relations among vehicles, as well as relations between vehicles and the whole scenario. As a most related work, \cite{GHOST} combines predicted visual abstractions and scalar confidence values to capture the perceptual uncertainty. However, they still rely on scalar confidence values produced by the independent perception system, which cannot be generalized to various interference and complex scenarios. 
In our framework, we do not explicitly identify ghost vehicles, instead, we learn a confidence representation, which implicitly evaluates the anomaly degree for each vehicle, to encourage the navigation model to learn how to effectively aggregate the information with confidence representations by itself, thus achieving end-to-end robust navigation. 
Our main contributions include:

Firstly, to the best of our knowledge, this is the first work to achieve end-to-end robust navigation against false positive detection. We introduce the relation learning to learn reliability representations of the surrounding vehicles' information and develop a robust navigation framework which can effectively aggregate all the information of normal and ghost vehicles to realize the safe navigation.

Secondly, in order to learn an efficient reliability representation, Graph Neural Networks (GNNs) are utilized to explicitly learn the spatio-temporal relation/consistency among vehicles' behaviors as well as between vehicles and the whole scenario. With end-to-end training paradigm, our approach fully exploits the potential of GNNs to encourage the network itself to learn how to represent the reliability and how to aggregate all the information with uncertainties for efficient navigation, thus increasing the efficiency and generalizability.

Finally, simulation results on the CARLA platform validate our effectiveness in the presence of ghost vehicles and our scalability to various dynamic scenarios.

\section{RELATED WORK}


\subsection{Learning Based Navigation}
Traditional methods decompose the complex navigation task into simpler sub-tasks\cite{pathPlanning} including environmental perception, localization, planning, and motion control. 
Recently, end-to-end navigation paradigm has attracted more and more attentions. 
Conditional imitation learning \cite{CIL} is the most widely used framework, which learns to map raw sensor data, such as LIDAR data \cite{imiLidar1} and image data \cite{cnnNavi1}, to output control signals directly. To improve the generalization performance, semantic information (such as BEVs) \cite{E2E1_semantic, E2E3_semantic} is widely used as input of the end-to-end framework. 



\subsection{Data Corruption in Self-Driving} 
Perception data corruption results from environmental disturbance/interference is inevitable in self-driving applications. 
In \cite{OOD}, a Generative Adversarial Network (GAN) is presented to detect out-of-distribution data. Degradation of visual data \cite{imageDinRain1} and LiDAR point cloud \cite{lidarDinRain} in various weather conditions also have been studied. As an example, an anomaly detection model is presented in \cite{lidarDinRain} to quantify LiDAR degradation in dynamic urban environments. 
Apart from the above natural perturbations, recent researches have shown that neural networks are very vulnerable to manually crafted adversarial disturbances \cite{PGD}, which are able to fool learning models with minimum disturbances that are indistinguishable to human. 
The authors in \cite{lidarAtt1} design a spoofing algorithm that misleads the state-of-the-art LiDAR perception system to observe fake objects in front of a victim vehicle, thus maliciously altering driving decisions. 
Arbitrary types of mis-detection can be achieved by introducing the targeted attacks on either the LiDAR data \cite{lidarAtt2, lidarAtt1} or visual data \cite{imgAtt1}. In addition, in order to make the disturbance more natural, GAN framework is used in \cite{aLittleFog} for data augmentation to generate adversarial images that are able to fool models for steering angle prediction in autonomous vehicles. 
As mentioned before, unreasonable braking incidents due to mis-detection have caused great losses to Tesla.

\subsection{Robust Self-Driving}
Several driving datasets are collected to cover various challenging conditions \cite{dataDvrive1}. Training with synthetic data of different weather conditions can also improve the robustness \cite{train Weather1}. Robust optimization methods for hardening control systems against image corruptions and other unexpected domain shifts are presented in \cite{robustNav}, however, only the steering control is considered. In addition, the robustness is achieved by adversarial training which is only effective to specific attacks. A soft BEV representation is presented in \cite{GHOST} to learn that if the information is trustworthy, thus avoiding excessively conservative behaviors in the presence of perception errors. However, they highly rely on scalar confidence values produced by an independent perceptual process, which can be totally fooled by adversarial attacks and produce a totally wrong confidence values. With end-to-end paradigm, our method does not explicitly identify the ghost vehicle, instead, we learn the confidence representation which implicitly contains anomaly evaluation for each vehicle. In this way, our model achieves end-to-end robust navigation by integrating all the information with uncertainties. 

\subsection{GNNs in Autonomous Navigation}
As an efficient way to learn non-Euclid data, 
GNNs are used to learn relations among surrounding vehicles for trajectory tracking and decision making tasks. To name a few, a spectral temporal GNN is proposed in \cite{GNN3} to capture inter-vehicle correlations to predict future trajectories. Authors in \cite{GNN2} utilize 
the spatio-temporal graph auto-encoder to recognise abnormal driving behaviors. 
In this paper, we propose the first work which utilizes GNNs to learn spatio-temporal consistency among vehicles in order to achieve robust navigation against corrupted data with ghost vehicles.

\section{Preliminaries}
{\it Ghost vehicle.} 
As mentioned before, we define the ghost vehicle as the false positive detection (which is a major mis-detection type in real cases), i.e.,  a vehicle that doesn't actually exist or appears in a wrong location. 

\begin{figure}[!t]
\centering
{\includegraphics[width=0.9\columnwidth]{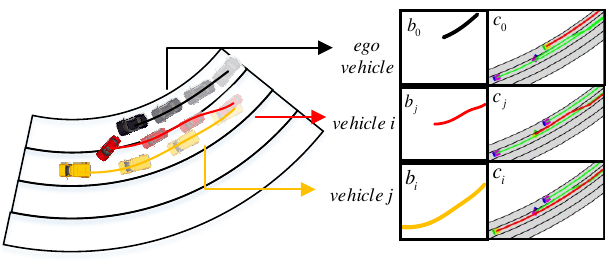}}
\caption{Input description. The left figure shows the overall scenario that contains the ego-vehicle (black), a ghost vehicle (red), and a normal vehicle (yellow). The right figure demonstrates the input of each vehicle, which contains historical trajectory information and corresponding BEV image. It should be noted that the model does not know which vehicle is the ghost vehicle, so every surrounding vehicles will be treated equally.}
\label{fig:input}
\end{figure}

{\it System inputs.} Our system input consists of three parts $\{\mathcal{B},\mathcal{C},\mathcal{N}\}$, where $\mathcal{B}$ is the historical trajectory information of all vehicles in a local observation range, $\mathcal{C}$ is the scenarios information and $\mathcal{N}$ is the desired route information of ego-vehicle. Similar to \cite{E2E1_semantic,E2E3_semantic}, the BEV information is considered. More specifically, we consider the historical trajectory information of $n$ vehicles surround the ego vehicle, i,e, $\mathcal{B}=\{b_{0}, b_{1}, ..., b_{n}\}$, where $b_{i}=\{o_{i}^{t-h+1}, o_{i}^{t-h+2}, ..., o_{i}^{t}\}$ denotes the $h$ temporal states of the $i^{th}$ vehicle. 
The $o_{i}^{t}=\{x_{i}^{t}, y_{i}^{t}, vx_{i}^{t}, vy_{i}^{t}, \theta_{i}^{t}\}$, 
where $x_{i}^{t}$ and $y_{i}^{t}$ denote the relative position to the ego vehicle in lateral and longitudinal directions respectively, $vx_{i}^{t}$ and $vy_{i}^{t}$ denote the relative velocity information, and $\theta_{i}^{t}$ denotes the relative heading. 
As shown in Fig.~\ref{fig:input}, we define $\mathcal{C}=\{c_{0}, c_{1}, ..., c_{n}\}$, where $c_{i}$ denotes the BEV map centered on the ego-vehicle, in which the trajectory of $i^{th}$ vehicle is marked with the red color. $\mathcal{N}$ is the desired navigation route of the ego-vehicle, which is represented by eight navigation waypoints. Note that the index $0$ represents the ego-vehicle and the index $1,...,n$ represents the surrounding vehicles (which may include the unknown ghost vehicle).


\begin{figure*}[!t]
\centering
{\includegraphics[width=2\columnwidth]{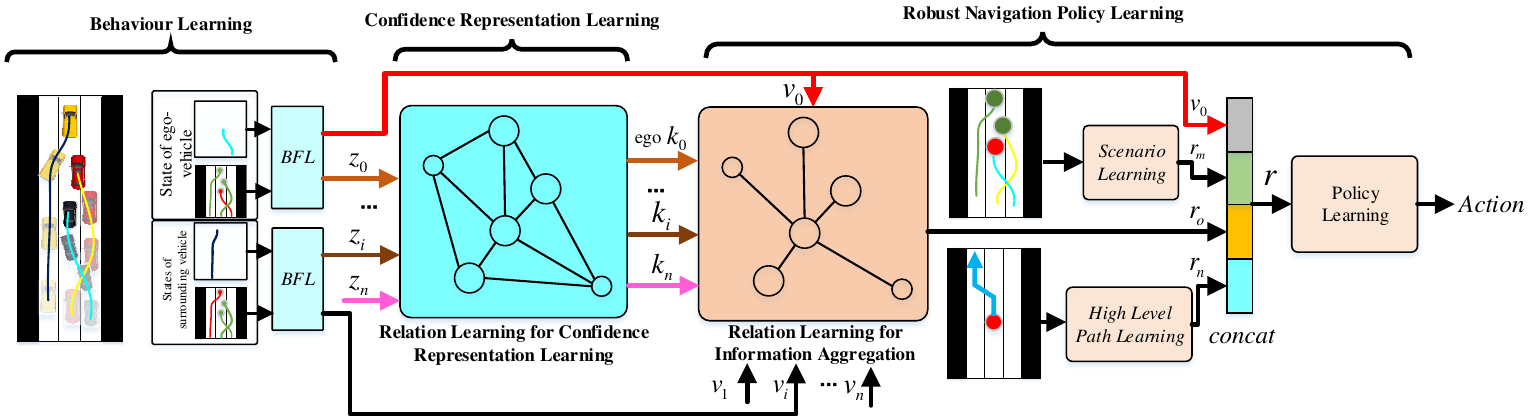}}
\caption{The overall framework of our proposed RLRN approach. In Behaviour learning module, the behaviour features $z_i$ is learned for each vehicle, which contains the information of  historical trajectory and the scenarios, in order to enable the anomaly information from the single vehicle aspect. Then, in Confidence Representation Learning module, the inconsistency among vehicles is learned through a GNN and the confidence representation $k_i$ is generated for each vehicle. After that, in Robust Navigation Learning module, information of all vehicles is aggregated to generate the overall relation features $r_o$, in addition, the scenario features $r_m$ and desired route features $r_n$ are also extracted. Finally, all the above features are input into the policy learning module to generate the final control actions.}
\label{fig:framework}
\end{figure*}

\section{Main Approach}

\subsection{System Framework}
We assume that, for each time step, the perception system may be totally fooled by disturbances or adversarial interference, and may detect one or more ghost vehicles, which can be generated at arbitrary positions. The ego-vehicle does not know whether there is a ghost vehicle and which neighbor is the ghost. 
As shown in Fig.~\ref{fig:framework}, our approach consists of three components:

{\it Behaviour Learning}: For each vehicle (including the ego-vehicle and the surrounding vehicles), we develop the behaviour feature learning (BFL) module to extract features 
$v_{i}$ from its historical trajectory information $b_i$ and also extract features $m_i$ from 
the BEV map $c_i$. 
Then we learn the behavior features $z_{i}$ by aggregating $v_{i}$ and $m_{i}$. 

{\it Confidence Representation Learning}: The behaviour features $z_i$ of all the vehicles are aggregated through a GNN for relation learning, including the consistency learning between the trajectory and scenario information of each vehicle, and the consistency learning among the information of all the vehicles. 
After that, the reliability representation vector ${k}_{i}$ of each vehicle, which implicitly contains the confidence evaluation for each vehicle, is produced for policy generation. 

{\it Robust Navigation Policy Learning}: We introduce residual concatenation to combine trajectory features $v_{i}$ with confidence representation features $k_{i}$ to generate the input of each vehicle, and then define another relation learning graph to aggregate all the information with uncertainties and output the overall relation feature $r_{o}$. In addition, a scenario learning module is designed to extract the whole traffic scenario features $r_{m}$, and the desired route features $r_{n}$ is also learned from $\mathcal{N}$. 
Finally, we combine 
$v_{0}$, $r_{m}$, $r_{n}$ and $r_{o}$ as the input of the final policy learning model and implement the imitation learning strategy to learn the final vehicle control signal ${a}$.

Our framework does not execute a ghost detection task explicitly, but implicitly learns a confidence representation vector $k_i$ to evaluate the reliability degree of each vehicle. In the training, Behaviour Learning, Confidence Representation Learning and Scenario Aggregation modules are coupled in the end-to-end paradigm, which enables the network to learn features that contain more efficient information, 
and learn how to use the learned features more reasonably. 





\begin{figure}[!t]
\centering
{\includegraphics[width=0.8\columnwidth]{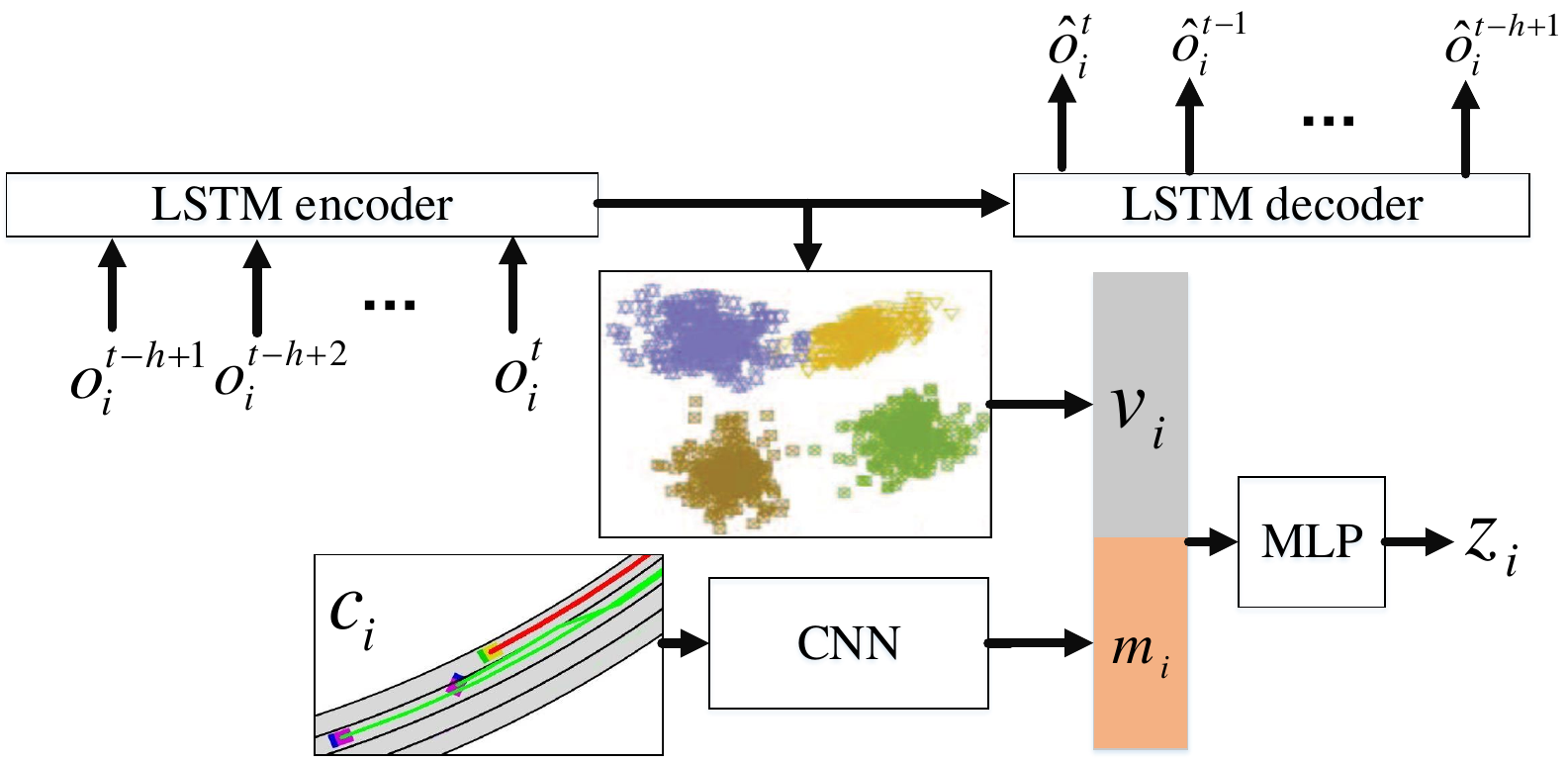}}
\caption{The detailed structure of Behaviour Feature Learning (BFL) model.}
\label{fig:Behaviour Feature Layer}
\end{figure}

\subsection{Behaviour Learning}
As shown in Fig.~\ref{fig:Behaviour Feature Layer}, the BFL module consists of LSTM-AutoEncoder and CNN based feature extractor. The LSTM-AutoEncoder is designed to extract efficient feature representations from the historical trajectory information of each vehicle. The LSTM-Encoder infers the sequential states in $b_i$ and learns a feature vector $v_{i}$, while the LSTM-Decoder reconstruct the sequential state $\hat{b}_i$. 
The LSTM-AutoEncoder is pre-trained with the reconstruction loss and will be fixed.
For the BEV map $c_i$ of each vehicle, the CNN backbone is adopted to learn the features $m_i$. 
Then we concatenate $v_i$ and $m_i$ and introduce an MLP to learn the behaviour features $z_i$ of each vehicle. 

\subsection{Confidence Representation Learning}

We introduce multi-head graph attention network (multi-head GAT) to generate the confidence representation vector $k_i$ of each vehicle through relation learning. A graph $G_{a}=\{V_{a}, E_{a}\}$ is designed, where $V_{a}$ includes all vehicles in a local observation range of the ego-vehicle, and $E_{a}$ is the adjacent matrix which is defined as $E_{a}=\{e_{ij}| 0\leq i, j \leq n\}$, where $e_{ij}=\frac{D-d_{ij}}{D}$ if the relative distance $d_{ij}\leq D$, otherwise, $e_{ij}=0$. In addition, we define $e_{ii}=1$. 

We define $z_i$ as initial node features.
During graph aggregation, the key functions $f_{k}$, query function $f_{q}$ and value function $f_{v}$ are constructed. The attention of vehicle $i$ to $j$ can be calculated as $\alpha_{ij} = softmax(f_{k}(z_{i})*f_{q}(z_{j}))*e_{ij}$. Finally, the confident representation vector can be generated as

\begin{equation}
    k_{i}=\mathop{\parallel} \limits_{w=1}^{W}\sum_{j}f^{w}_{v}(z_{j})*\alpha^{w}_{ij}.
\end{equation}
where W denotes the number of multi-head.

\subsection{Robust Navigation Policy Learning}


{\it Relation Learning.} As shown in Fig.~\ref{fig:framework}, we define another graph in which all vehicles will be only connected to the ego-vehicle. Then another multi-head GAT is introduced to learn the overall relation features $r_o$ only for the ego-vehicle. Inspired by \cite{c31} that inserts shortcut connections, we define the initial node features of each vehicle as $p_{i}=[k_{i},v_{i}]$.
The reason of re-introducing $v_i$ is that, 
in this work, $k_{i}$ is trained for confident representation, which may discard some information that is needed for navigation. Thus 
we make a residual concatenation as $p_{i}=[k_{i},v_{i}]$ and then aggregate $p_i$ of each vehicle through relation learning to produce the overall relation representation $r_{o}$. The process can be formulated as:
\begin{equation}
    r_{o}=\mathop{\parallel} \limits_{w=1}^{W}\sum_{j}f^{w}_{rv}(p_{j})*softmax(f^{w}_{rk}(p_{j})*f^{w}_{rq}(p_{0}))*e_{0j}
\end{equation}
where $f_{rv}$, $f_{rk}$, $f_{rq}$ is the value function, key function and query function.

{\it Action Generation}:
We pre-train a Variational Auto-Encoder (VAE) \cite{c32} to learn features $r_m$ from the whole traffic scenario $c_0$ with the reconstruction loss. 
$r_{m}$ contains necessary scenario information of the ego-vehicle for driving policy learning. In addition, the desired route features $r_n$ is extracted by using an MLP on the high level path information $\mathcal{N}$ of the ego-vehicle. Then we concatenate the ego-vehicle's behaviour features $v_{0}$, route features $r_n$, scenarios features $r_{m}$, and also the overall relation features $r_{o}$ as $r=[v_0,r_m,r_o,r_n]$. 
Finally, we utilize MLP networks to learn the final control signal ${a}$ through imitation learning.

\subsection{Training Process}
In order to ensure the network ability of our Confidence Representation Learning module, we first remove the Robust Navigation Policy Learning module and, instead, directly add 
an MLP for each vehicle as classification network to detect anomaly. The input of the MLP is $p_{i}=[k_i,v_i]$ and the output is a binary value to indicate if vehicle $i$ is a ghost vehicle. 
The model is pre-trained by solving the following problem: 
\begin{equation}
f*=\mathop{\arg\min}_{f} \sum l(f_{c}(V_{c}, E_{c}), y)
\end{equation}
where $y$ is the true label of the vehicle, $l$ is the cross-entropy loss between the class probability classification and the true label. We use these pre-trained parameters to initialize the weights of our Confidence Representation Learning module. 

Finally, the overall framework is trained end-to-end (with fixed parameters of LSTM-AutoEncoder and VAE modules). We use $L_2$-distance to calculate the imitation loss:
\begin{equation}
 L(a,a_{g})=\frac{1}{3}(|st-st_{g}|^2+|ac-ac_{g}|^2+|br-br_{g}|^2)
\end{equation}
where $st$, $ac$ and $br$ represent the steering angle, accelerator signal and brake signal, respectively, $g$ denotes the ground truth, i.e., the expert outputs.

\section{IMPLEMENTATION}

Our model is trained on one 3070 GPU with Pytorch framework, a computer with AMD Ryzen 7 5800X 8-Core CPU 32G RAM is used. The optimization optimizer is Adam. Xavier initialization is adopted to initialize the weights of the convolution layer. The shape of input RGB map is (800, 800, 3). The learning rate is set to be 0.001 with batch size 256. The maximum epoch is 50. We set $D=10m$ and $h=8$.


\subsection{Data Collection and Extrinsic Vehicle Injection}
CARLA's autopilots are utilized for data collection, 300 basic agents and 20 behaviour agents are generated in Town04 of the CARLA environment to produce the dataset. Basic agents serve as surrounding vehicles, while behaviour vehicles are used as the ego vehicle. During collection, the behaviour state of each ego-vehicle and its surrounding vehicles will be recorded. 
Besides, the navigation states of ego vehicles will be collected additionally, which contain control signal (steering, brake and throttle) and high level navigation path from CARLA. After that, ghost vehicles are randomly generated around the ego-vehicle. The initial speed and heading of ghost vehicles are generated according to a Gaussian distribution in which the mean value is defined as the average speed and orientation of all vehicles at each moment. The number of ghost vehicles varies from 0 to 2. The number of normal vehicles is fixed as 3 in each training sample, and varies from 1 to 6 in each evaluation sample. We collect 8000 training samples and 1000 evaluation samples for each combination of normal/ghost vehicle numbers.

\subsection{Comparison Models}

We train the following four models for comparison:
\begin{itemize}
    \item \texttt{Baseline}: Our full model that is trained with only the clean data (without any ghost vehicles).
    
    \item \texttt{RLRN-S}: Our full model that is trained with data samples which contain 3 normal vehicles and 1 ghost vehicle.
    
    \item \texttt{RLRN-T}: Our full network that is trained with 3 normal vehicles and various ghost vehicle numbers (which contains 0 to 2 ghost vehicles). 
    
    \item \texttt{CIL-Net}: The naive CIL framework \cite{CIL} is utilized to directly learn driving policy from $z_0$. This model is adversarial trained with data samples which contain 0-2 ghost vehicles.
\end{itemize}

\subsection{Metrics}
\begin{itemize}
    
    
    
    
    \item \texttt{Mean Accuracy (mAcc)}: To measure the regression accuracy, we use \texttt{mAcc} proposed in \cite{c29} to evaluate performance with different thresholds. With the threshold $\tau$, the value can be calculated as:
    \begin{equation}
    acc_{\tau}=\frac{1}{n} \sum_{i=1}^n {\bf 1}((|st-st_{g}|+|ac-ac_{g}|+|br-br_{g}|)<\tau)\nonumber
    \end{equation}
    where $n$ denotes the number of testing cases. Then 
    
    \begin{equation}
    mAcc=\frac{1}{|\Gamma|}\sum_{\tau} acc_{\tau \in \Gamma}
    \end{equation}
    where $\Gamma=\{0.2, 0.4, 0.8, 1.6\}$ is a set of empirically selected thresholds of control signals. 
    
    \item \texttt{Mean Corrupted Deviation (mCE)}: Similar to \cite{c29,Metric2}, \texttt{mCE} is adopted to evaluate the model's scalability by computing a weighted average of the error rate of a range of test data. The error rate of model $A$ on test set $h_{i}$ is defined as: $Err_{A}^{hi}=1-mAcc_{A}^{hi}$. 
    With the total test set $H=\{h_{1}, h_{2}, ..., h_{m}\}$, $mCE$ of the model $A$ can be calculated as
    \begin{equation}
    mCE_{A}=\frac{1}{m}\sum_{i=1}^m Err_{A}^{h_{i}}/Err_{baseline}^{h_{i}}
    \end{equation}
    By normalizing the error rate each model with that of the baseline model, $mCE_{A}$ is comparable across different conditions. 
    In this paper, we use datasets with different number of ghost vehicles (e.g. 0, 1 and 2) as test sets to evaluate the scalability to various cases. 

    \item \texttt{Mean Acceleration (mA)}: The average acceleration of all the cases: 
    \begin{equation}
    mA=\frac{1}{n}\sum_{i=1}^n ac_{i}
    \end{equation}
    
    \item \texttt{Mean Brake (mB)}: The average braking of all the cases 
    \begin{equation}
    mB=\frac{1}{n}\sum_{i=1}^n br_{i}
    \end{equation}
\end{itemize}

\section{EXPERIMENTAL RESULTS}

\begin{table}[t]
\renewcommand{\arraystretch}{1.2}
\caption{Robust navigation ability with various normal vehicles}
\label{Tab:result4}
\begin{center}
\renewcommand{\arraystretch}{1.2}
\scalebox{0.85}{
\begin{tabular}{c|c|c|c|c|c|c}
\hline
\hline
\multicolumn{1}{c|}{} & \multicolumn{5}{c}{$H=\{N_{g}|1,2\}$}\\
\cline{2-7}
\multicolumn{1}{c|}{} & \multicolumn{1}{c|}{$N_v=1$} & \multicolumn{1}{c|}{$N_v=2$} & \multicolumn{1}{c|}{$N_v=3$} & \multicolumn{1}{c|}{$N_v=4$} & \multicolumn{1}{c|}{$N_v=5$} & \multicolumn{1}{c}{$N_v=6$} \\
\cline{2-7}
\multicolumn{1}{c|}{} & ${mAcc}$ & ${mAcc}$ & ${mAcc}$ & ${mAcc}$ & ${mAcc}$ & ${mAcc}$ \\
\hline
\multicolumn{1}{c|}{\texttt{Baseline}} & 0.3133 & 0.4200 & 0.3256 & 0.3593 & 0.3203 & 0.5253 \\
\hline
\multicolumn{1}{c|}{\texttt{CIL-Net}} & 0.3577 & 0.4850 & 0.4603 & 0.4651 & 0.4790 & 0.3770 \\
\hline
\multicolumn{1}{c|}{\texttt{RLRN-S}} & 0.3743 & 0.4687 & 0.5220 & 0.5633 & 0.5736 & 0.6336 \\
\hline
\multicolumn{1}{c|}{\texttt{RLRN-T}} & \textbf{0.4536} & \textbf{0.5538} & \textbf{0.6212} & \textbf{0.6420} & \textbf{0.6691} & \textbf{0.6728}  \\
\hline
\hline
\end{tabular}}
\end{center}
\end{table}

\begin{table}[t]
\renewcommand{\arraystretch}{1.2}
\caption{Robustness scalability with various normal vehicles}
\label{Tab:result1}
\begin{center}
\renewcommand{\arraystretch}{1.2}
\scalebox{0.85}{
\begin{tabular}{c|c|c|c|c|c|c}
\hline
\hline
\multicolumn{1}{c|}{} & \multicolumn{5}{c}{$H=\{N_{g}|1,2\}$}\\
\cline{2-7}
\multicolumn{1}{c|}{} & \multicolumn{1}{c|}{$N_v=3$} & \multicolumn{1}{c|}{$N_v=1$} & \multicolumn{1}{c|}{$N_v=2$} & \multicolumn{1}{c|}{$N_v=4$} & \multicolumn{1}{c|}{$N_v=5$} & \multicolumn{1}{c}{$N_v=6$} \\
\cline{2-7}
\multicolumn{1}{c|}{} & ${mAcc}$ & ${mCE}$ & ${mCE}$ & ${mCE}$ & ${mCE}$ & ${mCE}$ \\
\hline
\multicolumn{1}{c|}{\texttt{Baseline}} & 0.3256 & 1 & 1 & 1 & 1 & 1 \\
\hline
\multicolumn{1}{c|}{\texttt{CIL-Net}} & 0.4603 & 0.9377 & 0.8882 & 0.8348 & 0.7663 & 1.3124 \\
\hline
\multicolumn{1}{c|}{\texttt{RLRN-S}} & 0.5220 & 0.9111 & 0.9161 & 0.6813 & 0.6272 & 0.7718 \\
\hline
\multicolumn{1}{c|}{\texttt{RLRN-T}} & \textbf{0.6212} & \textbf{0.7977} & \textbf{0.7697} & \textbf{0.5586} & \textbf{0.4868} & \textbf{0.6891}  \\
\hline
\hline
\end{tabular}}
\end{center}
\end{table}

\subsection{Scalability}

{\it Scalability to various number of normal vehicles}: Firstly, the scalability of each model 
to various normal vehicle numbers are tested. In this experiment, the number of ghost vehicles $N_g\in\{1,2\}$, while the number of normal vehicles $Nv\in\{1,2,3,4,5,6\}$. 
Please note that, in our training dataset, the number of normal vehicles is fixed as $Nv=3$. 
The results are demonstrated in the Tab.~\ref{Tab:result4} and Tab.~\ref{Tab:result1}. We can find that: 1) Comparing the four models, \texttt{RLRN-T} performs best among all the models, which has the strongest robustness (largest \texttt{mAcc}) and the highest scalability to unseen conditions (smallest \texttt{mCE}). 2) With the increasing number of normal vehicles, all models perform better. This result is reasonable as the more vehicles in the scenario will lead to the more cooperative driving behaviors, 
then the ghost vehicles can be learned easily through the relation learning, thus the ghost vehicles have less impact on the ego-vehicle. 3) Except from the \texttt{Baseline}, model which does not consider any ghost vehicles in training, \texttt{CIL-Net} performs worst among the rest models, even it is adversarial trained with ghost vehicles. This shows that, without the proposed relation learning, the network cannot achieve robustness to ghost vehicles. For 
\texttt{Baseline}, as it is never trained with ghost vehicles, thus being vulnerable to anomaly. 
4) Although \texttt{RLRN-S} is trained with the incomplete dataset (fixed nornal vehicle number, only one ghost vehicle), it still overwhelms \texttt{Baselien}, which demonstrates the advantage of our proposed RLRN framework. 

{\it Scalability to various number of ghost vehicles}: 
In this experiment, the number of normal vehicles $N_{v}\in\{3,4\}$, while we set the number of ghost vehicles as $N_{g}\in\{0,1,2\}$. The experimental results are shown in Table~\ref{Tab:result2}. The performance of \texttt{Baseline} is the best when there is no ghost vehicle. However, once the ghost vehicle occurs, its performance deteriorates rapidly. Besides, as the second best model in clean data, the performance of \texttt{RLRN-T} maintains consistent even one or more ghost vehicle occur, overwhelming all the rest models.

\subsection{Driving performance}
The presence of the ghost vehicles will lead to erratic and unreasonable driving behaviours of the ego-vehicle. The average accelerator and brake frequency are shown in Table~\ref{Tab:result3}. We can find that the existence of ghost vehicles significantly reduces the throttles of \texttt{Baseline} and \texttt{CIL-Net}, while increases their brake frequency. These results clearly show that the ghost vehicles have greatly affected their driving performance. In practice, the low throttle may lead to traffic jam, while unreasonable emergency brake can cause accidents. However, the presence of ghost vehicles has only a small effect on the output of \texttt{RLRN-S} and \texttt{RLRN-T}, which demonstrate our robustness. 

\begin{table}[t]
\renewcommand{\arraystretch}{1.2}
\caption{Experiment results with various abnormal vehicles}
\label{Tab:result2}
\begin{center}
\renewcommand{\arraystretch}{1.2}
\scalebox{0.85}{
\begin{tabular}{c|c|c|c|c|c|c}
\hline
\hline
\multicolumn{1}{c|}{} & \multicolumn{3}{c|}{$N_{v}=3$}& \multicolumn{3}{c}{$N_{v}=4$}\\
\cline{2-7}
\multicolumn{1}{c|}{} & \multicolumn{1}{c|}{$N_g=0$} & \multicolumn{1}{c|}{$N_g=1$}& \multicolumn{1}{c|}{$N_g=2$} & \multicolumn{1}{c|}{$N_g=0$} & \multicolumn{1}{c|}{$N_g=1$} & \multicolumn{1}{c}{$N_g=2$}\\
\cline{2-7}
\multicolumn{1}{c|}{} & ${mAcc}$ & ${mAcc}$ & ${mAcc}$ & ${mAcc}$ & ${mAcc}$ & ${mAcc}$ \\
\hline
\multicolumn{1}{c|}{\texttt{Baseline}} & \textbf{0.7167} & 0.3279 & 0.3233 & \textbf{0.7563} & 0.3581 & 0.3604 \\
\hline
\multicolumn{1}{c|}{\texttt{CIL-Net}} & 0.5040 & 0.5081 & 0.4127 & 0.5227 & 0.5227 & 0.4075 \\
\hline
\multicolumn{1}{c|}{\texttt{RLRN-S}} & 0.5163 & 0.5133 & 0.5307 & 0.5714 & 0.5693 & 0.5573 \\
\hline
\multicolumn{1}{c|}{\texttt{RLRN-T}} & 0.5720 & \textbf{0.6696} & \textbf{0.5727} & 0.5953 & \textbf{0.6798} & \textbf{0.6041} \\
\hline
\hline
\end{tabular}}
\end{center}
\end{table}


\begin{table}[t]
\renewcommand{\arraystretch}{1.2}
\caption{Driving performance evaluations}
\label{Tab:result3}
\begin{center}
\renewcommand{\arraystretch}{1.2}
\begin{tabular}{c|c|c|c|c}
\hline
\hline
\multicolumn{1}{c|}{} & \multicolumn{4}{c}{$N_{v}=3$}\\
\cline{2-5}
\multicolumn{1}{c|}{} & \multicolumn{2}{c|}{$N_{g}=0$} & \multicolumn{2}{c}{$N_g=1$}\\
\cline{2-5}
\multicolumn{1}{c|}{} & ${mA}$ & ${mB}$ & ${mA}$ & ${mB}$\\
\hline
\multicolumn{1}{c|}{\texttt{Baseline}} & 0.4481 & 0.1278 & 0.39 & 0.1406  \\
\hline
\multicolumn{1}{c|}{\texttt{CIL-Net}} & 0.5295 & 0.1625 & 0.4932 & 0.2052 \\
\hline
\multicolumn{1}{c|}{\texttt{RLRN-S}} & 0.4701 & 0.1268 & 0.4616 & 0.1257  \\
\hline
\multicolumn{1}{c|}{\texttt{RLRN-T}} & 0.4730 & 0.1248 & 0.4648 & 0.1223  \\
\hline
\hline
\end{tabular}
\end{center}
\end{table}

\subsection{Ablation Study}
We consider the following three variants for ablation study:
\begin{itemize}
    
    \item \texttt{RLRN$\backslash$.w C}: Our full network without the Confidence Representation Learning module.
    
    \item \texttt{RLRN$\backslash$.w R}: Our full network without residual concatenation of $v_{i}$. 
    In \texttt{RLRN$\backslash$.w K}, $k_{i}$ is directly taken as initial node features of the information aggregation graph in Robust Navigation Policy Learning module.
    
    \item \texttt{RLRN$\backslash$.w $f$}: Our full network that is trained with Confidence Representation Learning module being fixed by the initializing pre-trained parameters.
\end{itemize}

\begin{figure}[!t]
\centering
{\includegraphics[width=0.9\columnwidth]{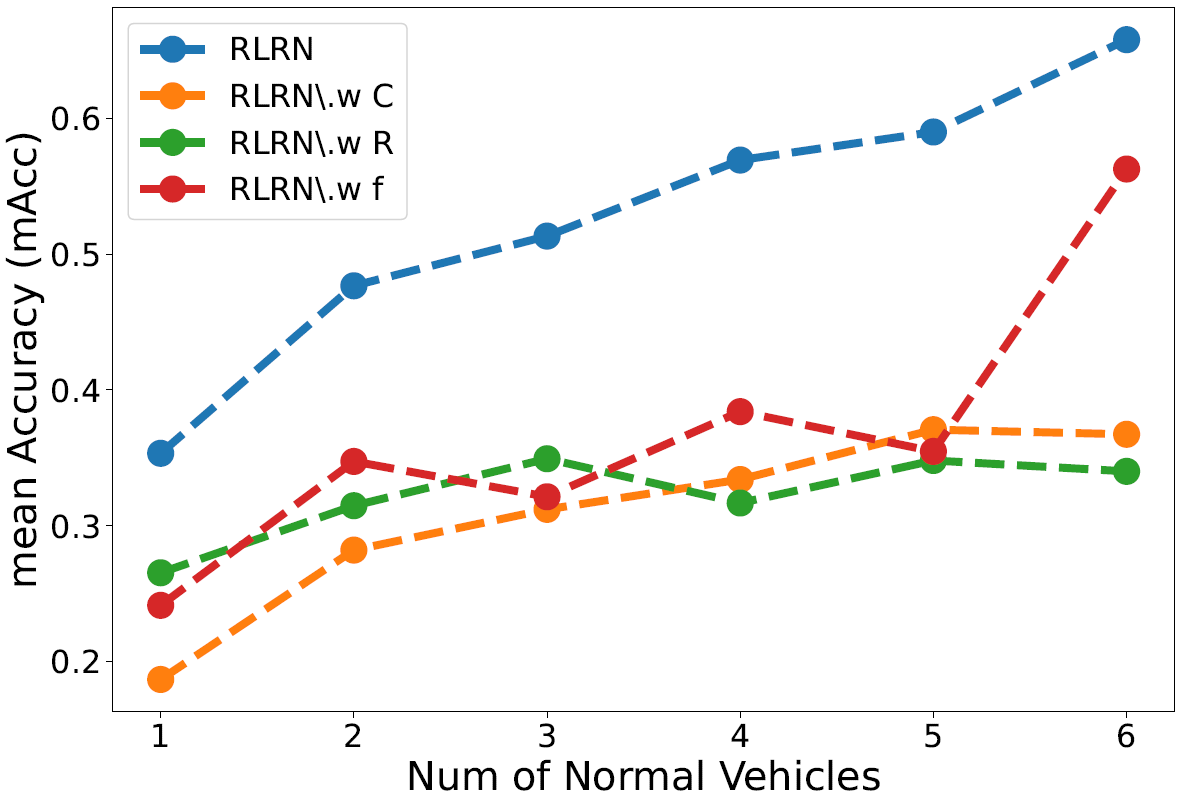}}
\caption{Ablation study results.}
\label{fig:abalation}
\end{figure}

The results of ablations studies are demonstrated in Fig.~\ref{fig:abalation}, which shows the change of mAcc with the increasing number of normal vehicles. 
We can observe that, once the Confidence Representation Learning module is removed in $RLRN\backslash.w C$, the performance decreases significantly, which validates that the presented confidence learning module is of great importance to the robustness to ghost vehicles. Comparing $RLRN$ and $RLRN\backslash.w R$, we can clearly see the effectiveness of concatenating $k_i$ and $v_i$. 
Besides, when the parameters in the Confidence Representation Learning module are fixed in $RLRN\backslash.w f$, the performance decreases largely, especially in the cases of small number of normal vehicles. 
These results show that the end-to-end paradigm encourages the model to learn more effective confidence representations and enables the model itself to learn how to aggregate all the vehicle information with uncertainties for efficient navigation. 

\section{CONCLUSION AND DISCUSSION}
In this paper, we provide the first work to achieve end-to-end robust navigation against ghost vehicles. We introduce the relation learning to learn reliability representations of the surrounding vehicles' information and develop a robust navigation framework which can effectively aggregate all the information of normal and ghost vehicles to realize the safe navigation. Simulation results on the CARLA platform validate our effectiveness and scalability to various number of ghost vehicles and normal vehicles.

In the future work, we plan to compare our approach with the methods which explicitly detect ghost vehicles from the perception aspect and conduct real experiments. 

\addtolength{\textheight}{-12cm}   









\end{document}